\documentclass[letterpaper, 10 pt, conference]{ieeeconf}  % Comment this line out if you need a4paper

\IEEEoverridecommandlockouts                              % This command is only needed if 
\usepackage{epsfig} % for postscript graphics files
\usepackage{amsmath} % assumes amsmath package installed
\usepackage{amssymb}  % assumes amsmath package installed
\usepackage{url}
\usepackage{booktabs} % professional-quality tables
\usepackage{cite}
\usepackage{subcaption}
\usepackage{caption}
\usepackage{graphicx}
\usepackage{xcolor}         % colors
\usepackage{comment}

\newcommand{\cut}[1]{}

\newcommand{\para}[1]{{\noindent\textbf{#1}}}
\newcommand{\parai}[1]{{\noindent\textit{#1}}}

\title{\LARGE \bf Transferring Implicit Knowledge of Non-Visual Object Properties Across Heterogeneous Robot Morphologies}

\author{
    \authorblockN{Gyan Tatiya$^{1}$ \quad Jonathan Francis$^{2}$  \quad Jivko Sinapov$^{1}$}
    \thanks{$^{1}$ Department of Computer Science, Tufts University, Email: {\tt\small \{Gyan.Tatiya, Jivko.Sinapov\}@tufts.edu}.
    $^{2}$ Bosch Center for AI, Email:
    {\tt\small Jon.Francis@us.bosch.com}}
}

\begin{document}

\maketitle
\thispagestyle{empty}
\pagestyle{empty}

%%%%%%%%%%%%%%%%%%%%%%%%%%%%%%%%%%%%%%%%%%%%%%%%%%%%%%%%%%%%%%%%%%%%%%%%%%%%%%%%
\begin{abstract}

%Humans interact with objects and observe several sensory modalities to develop intuition behind object properties.
Humans leverage multiple sensor modalities when interacting with objects and discovering their intrinsic properties.
Using the visual modality alone is insufficient for deriving intuition behind object properties (e.g., which of two boxes is heavier), making it essential to consider non-visual modalities as well, such as the tactile and auditory. 
%Vision modality alone is insufficient to derive all possible object properties making it essential to consider non-visual modalities such as audio and haptic.
%For example, using vision alone, we cannot find which one among two boxes is heavier.
Whereas robots may leverage various modalities to obtain object property understanding via learned exploratory interactions with objects (e.g., grasping, lifting, and shaking behaviors), challenges remain:
%by learning through exploratory interactions with objects (e.g., grasping, lifting, shaking), while processing multiple sensory modalities for learning interactive perception tasks.
%Robots can derive object properties by exploring objects (e.g., grasp, lift, shake) and processing sensory modalities for learning interactive perception tasks.
%However, k
the implicit knowledge acquired by one robot via object exploration cannot be directly leveraged by another robot with different morphology, because the sensor models, observed data distributions, and interaction capabilities are different across these different robot configurations.
%, because the distribution of data sensed by robots is different for each robot's morphology.
%Thus, each robot must learn the interactive perception task from scratch by exploring objects, which is a time-consuming process.
To avoid the costly process of learning interactive object perception tasks from scratch, we propose a multi-stage projection framework for each new robot for transferring implicit knowledge of object properties across heterogeneous robot morphologies.
% We propose two projection functions to transfer implicit knowledge about object properties, based on encoder-decoder networks and kernel manifold alignment.
% We collect a dataset using two heterogeneous robots that perform eight behaviors on 95 objects and evaluate our approach on two tasks: object-property recognition and object-identity recognition.
%We evaluate two projection functions within our framework to transfer implicit knowledge about object properties based on encoder-decoder networks and kernel manifold alignment.
% Our approach is able to learn the projection functions using the common object properties of different objects.
We evaluate our approach on the object-property recognition and object-identity recognition tasks, using a dataset containing two heterogeneous robots that perform 7,600 object interactions. % eight behaviors on 95 objects.
Results indicate that knowledge can be transferred across robots, such that a newly-deployed robot can bootstrap its recognition models without exhaustively exploring all objects.
We also propose a data augmentation technique and show that this technique improves the generalization of models.
We release code, datasets, and additional results, here: 
\footnotesize\url{https://github.com/gtatiya/Implicit-Knowledge-Transfer.}

\end{abstract}

%%%%%%%%%%%%%%%%%%%%%%%%%%%%%%%%%%%%%%%%%%%%%%%%%%%%%%%%%%%%%%%%%%%%%%%%%%%%%%%%
\section{Introduction}

% \begin{itemize}

% \item Why multi-sensory interaction is important for robots? Focus on non-visual modalities. \cite{tatiya2019sensorimotor, tatiya2020haptic, francis2022core}
% \item Why multi-sensory interaction is time consuming and hard to learn from scratch?
% \item How this research helps to solve the problem? Detecting properties of objects that are hidden from the visual sensor (e.g. weight, content)
% \item Main contributions:
% - Multi-robot, multi-sensory object interaction dataset.
% Question 1: property vs object based alignment for KEMA and Encoder-decoder.
% Question 2: object recognition vs property recognition.
% Question 3: Encoder-Decoder vs KEMA.

% \end{itemize}

Humans learn about object properties by physically interacting with objects and perceiving multiple sensory signals, including vision, audio, and touch \cite{thesen_Neuroimaging_2004, lederman1987hand, wilcox2007multisensory, ernst2004merging, gibson1988exploratory, chen2021framework}.
Interactions based on non-visual modalities such as audio and touch are essential, because vision alone is insufficient for detecting intrinsic object properties \cite{mccarthy1989artificial}: e.g., detecting whether an opaque bottle is full of liquid or empty. % would require processing the haptic sensation produced by the interaction.
Recent works show that learning implicit knowledge of non-visual object properties leads to robots' improved downstream performance, in material classification \cite{erickson2020multimodal}, liquid property estimation \cite{huang2022understanding}, object categorization \cite{tatiya2019deep}, and human-robot dialogue interaction \cite{thomason2017opportunistic}.
%Recent advances show robots can learn about non-visual object properties by interacting with objects for solving tasks such as material classification \cite{pmlr-v78-erickson17a}, liquid property estimation \cite{huang2022understanding}, and object categorization \cite{tatiya2019deep}.

% One way that a robot may learn about object properties is through performing exploratory interactions on objects and analyzing the effects via a diverse set of sensor modalities. The immediate issue is that this process is time-consuming, as it must be repeated for each additional robot. A natural desire may be to \textit{transfer} some representation of the object properties to a new robot, in order to enable it to learn faster and to complete its downstream tasks more efficiently. However, if the new robot has different interaction capabilities (e.g., different sensor models, or a different physical embodiment or \textit{morphology}), the implicit knowledge gained by the previous robot is not directly transferable to the present one. Indeed, a robot's machine learning model for such interactive perception tasks cannot be naturally applied to another robot, because these models are specific for each robot's embodiment, set of sensors, and its environment \cite{francis2022core}. While there is great need for transferring implicit knowledge of object properties across heterogeneous robot morphologies, to facilitate rapid learning and generalization, obtaining a general-purpose representation has remained challenging.

A robot may learn about object properties by performing exploratory interactions on objects and analyzing the effects via a diverse set of sensors \cite{malinovska2022connectionist, wei2021multimodal, liu2022texture}.
The immediate issue is that this process is time-consuming, as it must be repeated for each robot.
A natural desire may be to \textit{transfer} representation of the object properties to a new robot to enable it to learn faster and complete its downstream tasks more efficiently.
However, if the new robot has different interaction capabilities (e.g., different sensor models, or a different physical embodiment or \textit{morphology}), the implicit knowledge gained by the previous robot is not directly transferable to the new one.
Indeed, a robot's machine learning model for the interactive perception tasks cannot be naturally applied to another robot because these models are specific to each robot's embodiment, sensors, and environment \cite{francis2022core}.
While there is a great need to transfer implicit knowledge of object properties across heterogeneous robot morphologies, obtaining a general-purpose representation to facilitate rapid learning has remained challenging.

%One way to learn an object property representation is to have each robot interact with objects and collect their multi-sensory datasets, which is time-consuming.
%Furthermore, a robot's machine learning model for such interactive perception tasks cannot be directly applied to another robot because these models are specific for each robot's embodiment and set of sensors.

%Unfortunately, there is no general-purpose representation of non-visual object properties, given a robot sensor model and an embodiment. 

%In addition, there is no general-purpose knowledge representation for non-visual modalities of robots.

%As a result, there is a need to transfer the implicit knowledge about the object properties across heterogeneous robot morphologies to facilitate rapid learning.

To address this challenge, we propose a framework that leverages learned projection functions to transfer implicit knowledge of non-visual object properties from a more-experienced source robot to a newly-deployed target robot.
% Specifically, we consider the encoder-decoder network (EDN) and kernel manifold alignment (KEMA) methods for identifying object property-based and object identity-based correspondences, across the projections from the source and target robot models.
Specifically, we consider the general encoder-decoder network (EDN) model class \cite{badrinarayanan2017segnet} and the kernel manifold alignment (KEMA) method \cite{tatiya2020haptic, liu2018transferable, wang2011heterogeneous} as projection functions for learning object property-based and object identity-based correspondences.
%
%We evaluate two projection functions, encoder-decoder network (EDN) and kernel manifold alignment (KEMA), on two tasks, object-property, and object-identity recognition tasks.
%We evaluate two projection functions, encoder-decoder network (EDN) and kernel manifold alignment (KEMA), to build the source and target correspondences to learn the projection functions: object property-based and object identity-based correspondences.
To test our framework, we collected a dataset of two robots, {\it Baxter} and {\it UR5}, that performed eight behaviors on 95 objects.
We evaluate our framework on two tasks: object-property and object-identity recognition tasks.
% The results of our experiments show that in the object-property recognition task, transferring knowledge from robots to a shared latent space using KEMA boosts the performance of the target robot, especially when it has limited time to learn the task.
% Moreover, in the object-identity recognition task, using the projected features transferred from the source robot using EDN enables the target robot to perform better than a condition in which it learns using $100\%$ of its own features, indicating knowledge transfer accelerates the target robot's learning.
The results of our experiments show that KEMA learned using object identity-based correspondence consistently outperforms EDN in both tasks indicating transferring knowledge from robots to a shared latent space boosts the performance of the target robot.
Furthermore, we propose a data augmentation technique independent of the learning task and show that using our data augmentation technique improves the models' generalization and prevents overfitting.

%%%%%%%%%%%%%%%%%%%%%%%%%%%%%%%%%%%%%%%%%%%%%%%%%%%%%%%%%%%%%%%%%%%%%%%%%%%%%%%%
\section{Related Work}

%\para{Multi-sensory object interaction.}
%\para{Interactive perception of object properties.}
\para{Interactive object perception:}
% Explain some existing multi-sensory interaction based work. Why is is challenging?
Studies in psychology and cognitive science show that humans manipulate objects in multiple stages to extract information about their properties, such as texture, stiffness, temperature, and weight \cite{klatzky1992stages, lederman1993extracting, di2002role}.
In addition, the human brain leverages a multisensory representation when recognizing object properties, enabling flexible generalizability to unknown contexts \cite{lacey2007vision, lacey2014visuo}.
Recent advances in intelligent robotics consider integrating multisensory information acquired by object exploration \cite{bohg2017interactive, tatiya2019deep, pastor2020bayesian, sawhney2020playing, navarro2022visuo, li2020review, wang2022audio}, where one challenge is that the implicit knowledge acquired by one robot through interactive perception cannot be directly transferred to another robot: the unique nature of the robot's embodiment drastically affects the sensed data distribution and resultant model that each robot learns. % of the world. %the heterogeneous nature of robots severely impacts the distribution of sensed data, which affects the representation each robot uses to learn its machine learning model.
Whereas the focus of prior work has been limited to learning from scratch for each robot \cite{sinapov_grounding_2014, falco2019transfer, tatiya2019sensorimotor}, this is prohibitively expensive at scale, e.g., for a fleet of heterogeneous robots.
% via interactive perception tasks, 
%Consequently, prior work has focused on learning via interactive perception tasks from scratch, for each robot \cite{sinapov_grounding_2014, falco2019transfer, tatiya2019sensorimotor}, which would be prohibitively expensive at scale, e.g., for a large-scale deployment of heterogeneous robots.
%Due to this, each robot has to learn the interactive perception task from scratch by exploring objects, which is an expensive process.
%To tackle this limitation, w
We propose a framework for transferring implicit knowledge about object properties from a source robot to a target robot. %, for faster learning on the target. %To the best of our knowledge, we are the first to provide such a framework.% for transferring implicit knowledge of object properties, across heterogeneous robot morphologies.

%\para{Object property transfer.}
\para{Transferring knowledge of object properties:}
% Compare with previous works.
% \begin{itemize}
% \item Explain the ICDL paper's approach and its limitations.
% \item Explain IROS paper's approach and its limitations.
% \item How this paper address those limitations:
% - Real robots dataset.
% - Multiple property: weight, content.
% - property vs object based alignment, object vs property recognition, Encoder-Decoder vs KEMA.
% \end{itemize}
Recent work demonstrates that implicit knowledge from the interactive object perception can be transferred across sensor models and robots \cite{falco2019transfer, tatiya2019sensorimotor, tatiya2019deep, tatiya2020haptic, sinapov_grounding_2014, tatiya2020framework}. In \cite{sinapov_grounding_2014}, a robot performed interactive object perception to improve object category recognition. As implicit knowledge transfer was not the focus of that work, experiments were conducted on only a single robot. Moreover, whereas object properties may sometimes be the same for objects in different categories (e.g., bottles and cups can have similar colors, contents, and weights), their method encouraged unconstrained feature similarity based on object category alone, compromising prospects for transferring the features across robots or tasks. Our cross-robot transfer approach jointly learns to distinguish between different categories while leveraging learned similarities across properties. %Furthermore, under their task definition, object properties may be the same for objects in different categories (e.g., some bottles and cups have similar colors and weights); consequently, their method encouraged feature similarity for even different object categories, thereby compromising the prospects for transferring the learned features to a different robot or task.
%The range of experiments in this work were conducted on a single robot; furthermore, it is unclear what prospects there are for generalization based on the learned features, at the categorization described in the work, since their method encouraged feature similarity for even different object categories.
%%In \cite{falco2019transfer}, the authors consider transferring implicit knowledge across different sensor modalities (e.g., vision to haptic), although again within a single robot.
In \cite{tatiya2019sensorimotor}, authors consider object categorization under a transfer learning paradigm, %between a ``source robot'' and a ``target robot'', 
wherein an encoder-decoder network was used to generate a ``target'' robot's features from a ``source'' robot's learned representation. The authors use only a single robot in their experiments; however, %Because the authors had only one physical robot available for experiments, however, the source and target robots had the same embodiment, 
so inherent challenges introduced by different robot morphologies remain to be studied.
%A shared set of object categories in the dataset of \cite{sinapov_grounding_2014} was used to learn the mapping between the two robots' feature spaces. However, it was unclear what object properties can be captured in the target robot's generated features because some object properties might be the same for objects across different object categories. For example, some objects in the bottles and cups categories have similar colors and weights. Furthermore, due to the absence of multiple robots in the dataset, a physically identical robot was used that performed different behaviors as the source and target robots. Thus, there were minimal configuration changes across the source and target robots.
The approach in \cite{tatiya2020haptic} was used to project features from 3 robots with different embodiments to a shared latent space for object-identity recognition. However, their experiments consisted of only simulated robots that recorded only one sensor modality (effort) during interaction with objects that varied in only one dimension (weight).
To address these shortcomings, we collect a multisensory dataset using two real robots with different morphologies that explore 95 objects that vary by color, weight, and contents. %We make this dataset publicly available to foster research in knowledge transfer in robots. 
We develop a multi-stage projection method for implicit knowledge transfer across two heterogeneous robots, and we evaluate our approach on object-property recognition and object-identity recognition. %To find the best way to build correspondences using the EDN and KEMA for different tasks, we also evaluate object-property and object-identity-based correspondences.

% \begin{comment}
%\para{Object interaction datasets.}
\para{Interactive object perception datasets:}
% \jf{Consider adding comparisons with other datasets for, e.g., object category recognition -- see \cite{gao2022objectfolder}}
Compared to existing object interaction datasets \cite{sinapov_grounding_2014, tatiya2020haptic, Gao_2022_CVPR}, ours offers additional value for research needs.
In \cite{sinapov_grounding_2014}, the dataset only contained a single robot, whereas we collected our dataset using two robots with different morphologies.
In \cite{tatiya2020haptic}, simulated robots were used that collected only effort signals during object interaction. In contrast, we used real-world robots and collected multiple sensory signals, including vision, audio, and haptic.
In \cite{Gao_2022_CVPR}, the audio and tactile signals correspond to impact or touch behavior performed on 3D virtualized objects.
However, we collected the visual and non-visual sensory modalities while the robots performed several exploratory behaviors (e.g., grasp, shake) on 95 real-world objects that vary in multiple dimensions (color, weight, and content).
To the best of our knowledge, our dataset contains the largest number of objects, with the most dimensions of distinction ever explored by
multiple real robots for transferring implicit knowledge.
% To the best of our knowledge, our dataset contains the largest number of objects, with most dimensions of distinction, ever explored by multiple real robots for implicit knowledge transfer learning.
% \end{comment}

%%%%%%%%%%%%%%%%%%%%%%%%%%%%%%%%%%%%%%%%%%%%%%%%%%%%%%%%%%%%%%%%%%%%%%%%%%%%%%%%
\section{Learning Methodology}
\label{sec:approach}

\begin{figure*}[!ht]
\centering
\includegraphics[width=\textwidth]{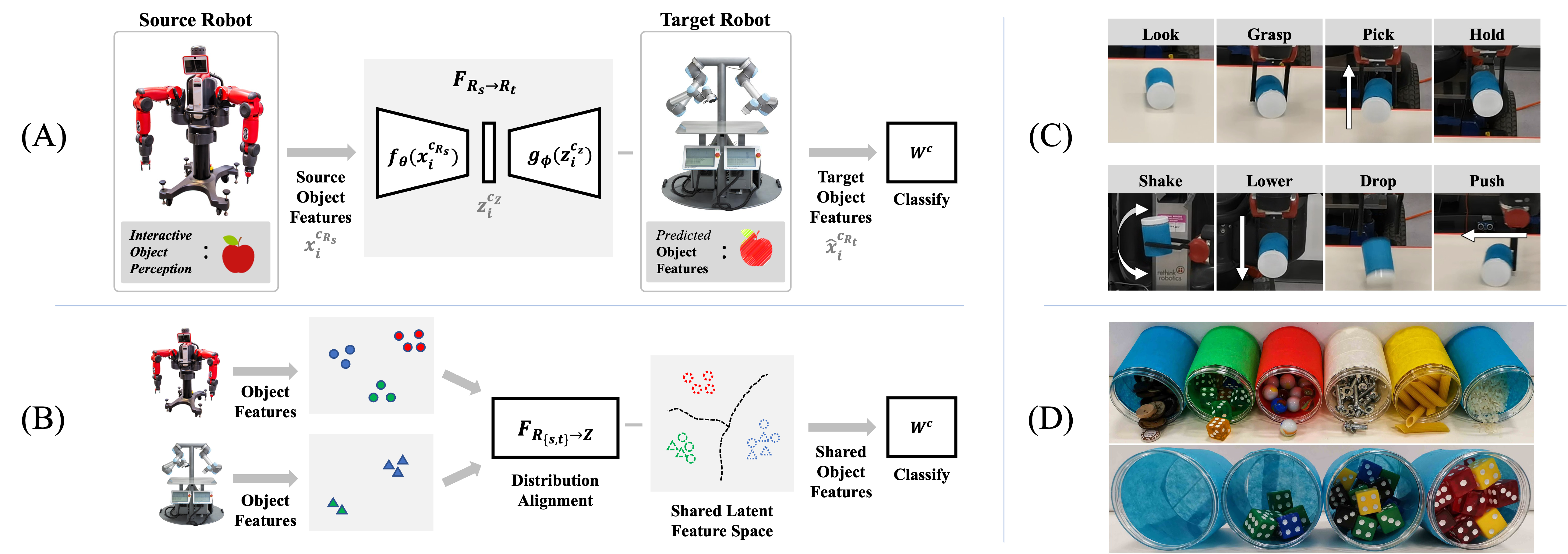}
\caption{\small (A) Shows projection from {\it Baxter} to {\it UR5} using Encoder-Decoder Network (EDN). 
(B) Shows projection from {\it Baxter} and {\it UR5} to a shared latent space using Kernel Manifold Alignment (KEMA). (C) The 8 exploratory behaviors used to learn about the objects.
(D) The 95 objects used in this study vary in: \textbf{(top)} colors (\textit{blue}, \textit{green}, \textit{red}, \textit{white}, and \textit{yellow}), contents ({\it wooden buttons}, {\it plastic dices}, {\it glass marbles}, {\it nuts} \& {\it bolts}, {\it pasta}, and {\it rice}), and \textbf{(bottom)} weights ({\it empty}, {\it 50g}, {\it 100g}, and {\it 150g}).}
\label{fig:System_Overview}
\vspace{-0.5cm}
\end{figure*}

\subsection{Notation and Problem Formulation}

% Let a robot interact with a set of objects $\mathcal{O}$ that vary in non-visual properties (e.g., {\it weight}, {\it sound}).
% The robot performs a set of exploratory behaviors $\mathcal{B}$ (e.g., {\it grasp}, {\it lift}) while recording sensory signals from a set of non-visual sensory modalities $\mathcal{M}$ (e.g., {\it audio}, {\it force}).
% Let $\mathcal{C}$ be the set of sensorimotor contexts, including each possible behavior $b \in \mathcal{B}$ and sensory modality $m \in \mathcal{M}$ combination (e.g., {\it grasp-audio}, {\it lift-force}).
% For a given context $c \in \mathcal{C}$, let the robot interact with each object for $n$ trials.
% Finally, let there be ${R}$ such robots with $\mathcal{B}_r$, $\mathcal{M}_r$, $\mathcal{C}_r$, and $n_r$, where $r = 1, ... {R}$.
% For each exploration trial, the robot $r$ observation feature is represented as $x_{c_r}^{i} \in \mathbb{R}^{D_{c_r}}, i = 1, ..., n_r$ where $D_{c_r}$ is the dimensionality of the features space for robot $r$ under context $c_r$.

Consider two robots with different morphologies, represented as source and target robots, or ${R}_s$ and ${R}_t$ respectively.
For a given robot ${R}$, let $\mathcal{B}_R$ be the set of exploratory behaviors (e.g., {\it grasp}, {\it lift}) and let $\mathcal{M}_R$ be the set of non-visual sensory modalities (e.g., {\it audio}, {\it force}).
Let $\mathcal{C}_R$ be the set of sensorimotor contexts, including each possible combination of a behavior in $\mathcal{B}_R$ and a sensory modality in $\mathcal{M}_R$ (e.g., {\it grasp-audio}, {\it lift-force}).
For an exploration trial, the robot $R$ performs exploratory behaviors $\mathcal{B}_R$ on a specific object and records a sensory signal for each modality in $\mathcal{M}_R$.
There are $n_{R}$ such exploration trials on each object.
For the $i^{th}$ exploration trial, robot ${R}$'s observation feature is $x^{c_{R}}_{i} \in \mathbb{R}^{D_{c_{R}}}$, where $i \in \{1, ..., n_{R}\}$, $c_{R} \in \mathcal{C}_R$, and $D_{c_{{R}}}$ is the dimension of robot $R$'s feature space under context $c_{R}$.

% Our main goal is to effectively learn a projection function to transfer knowledge gained by object interaction from a greater experienced source robot to a less experienced target robot.
% More formally, we divide our total set of possible objects $\mathcal{O}$ into two mutually exclusive subsets: $\mathcal{O}_{\text{shared}}$ and $\mathcal{O}_{\text{source-only}}$.
% Both target and source robots have interacted with objects in $\mathcal{O}_{\text{shared}}$ and only source robot has interacted with objects in $\mathcal{O}_{\text{source-only}}$.
% Thus, the source robot has experience with more objects.
% We learn the projection function using the objects in $\mathcal{O}_{\text{shared}}$ and transfer knowledge about the objects in $\mathcal{O}_{\text{source-only}}$ to the target robot using the learned projection function.
% This knowledge transfer will enable the target robot to learn the object property recognition task faster with fewer object interactions than the source robot.
% The projection function can project the observation feature from the source robot's feature space to the target robot's feature space or from each robot's feature space to a shared latent feature space.
% In the former case, we train the target robot in its own feature space, whereas, in latter case, we train the target robot in the latent space.

Let $\mathcal{O}$ be the set of objects that vary in non-visual properties (e.g., {\it weight}, {\it sound}).
We assume that the source robot has explored each object $n_{{R}_s}$ times, whereas the target robot has comparatively less experience.
More specifically, either the target robot has only explored a subset $\mathcal{O}_t \subset \mathcal{O}$ or explored an object for less trials than $n_{{R}_s}$ (i.e., $n_{{R}_t} < n_{{R}_s}$).
Our goal is to learn a projection function to transfer knowledge gained through object interaction, from the more-experienced \textit{source} robot to the less-experienced \textit{target} robot.
We learn the projection function using the common objects experienced by both robots and transfer knowledge about the source robot's additional experience by using the learned projection function.
This knowledge transfer will help the target robot learn about object properties faster, with fewer object interactions, and predict the properties of novel objects.

We consider learning two projection functions.
First, the projection function $F_{{R}_s \rightarrow {R}_t}$, that projects the observation features from the source robot's feature space to the target robot's feature space.
More specifically, $F_{{R}_s \rightarrow {R}_t}:x_i^{c_{{R}_s}} \rightarrow \hat{x_i}^{c_{{R}_t}}$, where $\hat{x_i}^{c_{{R}_t}}$ is the projected features in the target robot's feature space.
Second, the projection function $F_{{R} \rightarrow \mathcal{Z}}$, that projects the observation features from each robot's feature space to a shared latent feature space.
More specifically, $F_{{R}_s \rightarrow \mathcal{Z}}:x_i^{c_{{R}_s}} \rightarrow z_i^{c_{\mathcal{Z}}}$ and $F_{{R}_t \rightarrow \mathcal{Z}}:x_i^{c_{{R}_t}} \rightarrow z_i^{c_{\mathcal{Z}}}$, where $z_i^{c_{\mathcal{Z}}} \in \mathbb{R}^{D_{\mathcal{Z}}}$ and represents the shared latent features of size $D_{\mathcal{Z}}$.
In the first mapping, we train the target robot in its own feature space; for the second mapping, we train the target robot in the shared latent space.

We also consider two ways to build correspondences between the source and the target robots, for learning the projection functions. 
First, object-identity correspondence, in which the source-target pair corresponds to the same object identity.
It is applicable when both robots have access to the same objects.
Second, object-property correspondence, in which the source-target pair corresponds to the same object property.
It is applicable when both robots operate in different environments and do not have access to identical objects but have access to objects with the same properties (e.g., red and blue bowls containing rice).

\subsection{Projection to Target Feature Space}

% The first projection is responsible for performing density estimation, to characterise the likelihood that source and target robot features result from similar observations; here, we propose the use of the standard encoder-decoder modeling class, which may be realised by popular architectures (e.g., transformers, dense convolutions, or MLPs), depending on the nature of the data presentation.

We propose using an Encoder-Decoder Network (EDN) \cite{tatiya2019sensorimotor} to train the projection function $F_{{R}_s \rightarrow {R}_t}$, mapping observation features from the source robot's feature space to the target robot's feature space (Fig. \ref{fig:System_Overview}A).
First, encoder $f_{\theta}$ transforms the observation feature of the source robot $x_i^{c_{{R}_s}}$ into a fixed-size lower-dimensional vector $z_i^{c_{\mathcal{Z}}} \in \mathbb{R}^{D_z}$ of size $D_z$.
Then, decoder $g_{\phi}$ uses this code vector $z_i^{c_{\mathcal{Z}}}$ to generate the predicted observation feature of the target robot $\hat{x_i}^{c_{{R}_t}}$.
We denote this overall non-linear mapping as $F_{{R}_s \rightarrow {R}_t}$: $\hat{x_i}^{c_{{R}_t}} = g_{\phi}(f_{\theta}(x_i^{c_{{R}_s}}))$, where $\theta$ and $\phi$ are network parameter weights of encoder and decoder, respectively. For training the EDN, we use a dataset of source-target feature pairs $\{x_i^{c_{{R}_s}}, x_{i}^{c_{{R}_t}}\}_{i=1}^N$, with $N$ training samples.
We optimize EDN parameters by minimizing root mean-squared error (RMSE) between real features observed by target robot ${x_i}^{c_{{R}_t}}$ and ``generated'' target features $\hat{x_i}^{c_{{R}_t}}$ obtained by applying the projection to the corresponding source features:  $\theta^{\star}, \phi^{\star}=
\underset{\theta, \phi}{\arg\min}
\sqrt{\frac{1}{N} \sum_{i=1}^{N} (x_{i}^{c_{{R}_t}} - {\hat{x}_{i}^{c_{{R}_t}}})^2}$. %
\begin{comment}
{\color{red} We use root mean square error (RMSE) to compute the error and minimize:
\begin{equation}
\theta^{\star}, \phi^{\star}=
\underset{\theta, \phi}{\arg\min}
\sqrt{\frac{1}{N} \sum_{i=1}^{N} (x_{i}^{c_{{R}_t}} - {\hat{x}_{i}^{c_{{R}_t}}})^2}
\label{loss_function}
\end{equation}}
\end{comment}
Given a trained EDN, we generate the target robot's feature to transfer knowledge about the source robot's additional experience; then, using a standard multi-class classifier, we can train the target robot to recognize object properties with the ``generated'' features.

\begin{comment}
\begin{figure}
\centering
%\includegraphics[width=7cm]{figs/System_Overview.jpg}
\includegraphics[width=\linewidth]{figs/projections.png}
\caption{\small (a) Shows projection from {\it Baxter} to {\it UR5} using Encoder-Decoder Network (EDN). (b) Projection from {\it Baxter} and {\it UR5} to a shared latent space using Kernel Manifold Alignment (KEMA).}
\label{fig:System_Overview}
\end{figure}
\end{comment}

\subsection{Projection to Shared Latent Feature Space}

The projection $F_{{R} \rightarrow \mathcal{Z}}$ can be achieved through distribution alignment---organizing observation features from each robot's feature space within a shared representation (Fig. \ref{fig:System_Overview}B). We illustrate this mapping via Kernel Manifold Alignment (KEMA) \cite{tatiya2020haptic}, which constructs a set of domain-specific projection functions for each robot $F_{{R}} = [F_{{R}_s}, F_{{R}_t}]^T$, such that the examples of the same object property would locate closer while examples of different object properties would locate distantly. To compute the data projection matrix $F_{{R}}$, we minimize the cost related to the projection functions being too dissimilar: $\{F_{{R}_s}, F_{{R}_t}\} = \underset{F_{{R}_s}, F_{{R}_t}}{\arg\min}(C(F_{{R}_s}, F_{{R}_t}))$. % We propose using Kernel Manifold Alignment (KEMA) \cite{tatiya2020haptic} to train the projection function $F_{{R} \rightarrow \mathcal{Z}}$, for mapping observation features from each robot’s feature space to a shared feature space (Fig. \ref{fig:System_Overview}b). KEMA constructs a set of domain-specific projection functions, $F_{{R}} = [F_{{R}_s}, F_{{R}_t}]^T$, that project data from source and target robots into a common latent space such that the examples of the same object property would locate closer while examples of different object properties would locate distantly. To compute the data projection matrix $F_{{R}}$, we minimize the cost related to the projection functions being too dissimilar: $\{F_{{R}_s}, F_{{R}_t}\} = \underset{F_{{R}_s}, F_{{R}_t}}{\arg\min}(C(F_{{R}_s}, F_{{R}_t}))$. %
\begin{comment}
\begin{equation}
\begin{split}
    & \{F_{{R}_s}, F_{{R}_t}\} = \underset{F_{{R}_s}, F_{{R}_t}}{\arg\min}(C(F_{{R}_s}, F_{{R}_t}))
\label{eq:cost_function}
\end{split}
\end{equation} %
\end{comment}
Here, $C(\cdot) = \frac{1}{\texttt{DIS}} (\mu * \texttt{GEO} + (1-\mu) * \texttt{SIM})$, where the geometry of a domain, class similarity, and class dissimilarity are represented as \texttt{GEO}, \texttt{SIM}, and \texttt{DIS}, respectively.
\texttt{GEO} is minimized to preserve the local geometry of each domain by penalizing projections in the input domain that are far from each other.
\texttt{SIM} is minimized to encourage examples with the same object property to be located close to each other in the latent space by penalizing projections of the same object property mapped far from each other.
\texttt{DIS} is maximized to encourage examples with different object properties to be located far apart in the latent space by penalizing projections of the different object properties that are close to each other.
The parameter $\mu \in [0, 1]$ regulates the contribution of the \texttt{GEO} and the \texttt{SIM} terms.
For more details on KEMA, please see \cite{tuia_kernel_2016}. Data in the latent feature space are comparable and can be used to train a standard multi-class classifier for different robots.
The target robot can use this classifier to recognize properties of objects it has never interacted with.%, as long as the source robot has interacted with them.

\subsection{Model Implementation and Training}%\footnote{Datasets and source code for study replication are available at: \url{https://github.com/gtatiya/Implicit-Knowledge-Transfer}}}

Specific EDN architectures (e.g., transformers, dense convolutions, etc.) may be chosen according to the form of the data observations; in our experiments, we used an architecture that consists of three fully-connected layers for both encoder and decoder, with 1000, 500, 250 units, activation via Exponential Linear Units (ELU), and a 125-dimensional latent code vector. We use Adam \cite{kingma_adam_2015} with a learning rate of $10^{-4}$ to compute gradients according to RMSE, over 1000 epochs.
We used Radial Basis Function (RBF) for KEMA's kernel function, with $\mu = 0.5$.
We train the target robot's recognition model via a multi-class SVM with the RBF kernel.
For the EDN approach, this recognition model is trained using the ``generated'' features from the source robot and the real features of the target robot used to train the EDN; for the KEMA approach, this recognition model is trained using the shared latent features corresponding to both robots' datapoints used to learn the KEMA projection function.

%%%%%%%%%%%%%%%%%%%%%%%%%%%%%%%%%%%%%%%%%%%%%%%%%%%%%%%%%%%%%%%%%%%%%%%%%%%%%%%%
\section{Evaluation}

\subsection{Experimental Platform and Feature Extraction}

\subsubsection{Robots and Sensors}

% We collected our dataset using two robots: {\it Baxter} \cite{baxter_site, baxter_fig} and {\it UR5} \cite{ur5_site, ur5_fig} (shown in Fig. \ref{fig:System_Overview}A).
We collected our dataset using two robots: {\it Baxter} \cite{baxter_fig} and {\it UR5} \cite{ur5_fig} (Fig. \ref{fig:System_Overview}A).
{\it Baxter} has dual 7-degree-of-freedom (DOF) arms and a 2-finger gripper.
We used the left {\it Baxter} arm for the data collection.
{\it UR5} has 6-DOF and 2-finger Robotiq 85 gripper.
%Table \ref{tab:modalities} shows the list of both robots' modalities and their sampling rates.
{\it Baxter} had a PrimeSense camera %\footnote{\url{https://en.wikipedia.org/wiki/PrimeSense}} 
mounted on its head, which captures 640$\times$480 images, and an Audio-Technica PRO 44 microphone %\footnote{\url{https://www.amazon.com/dp/B0002BBOOS}} 
placed on its workstation.
{\it Baxter} hand camera captures 480$\times$300 images.
{\it UR5} had an Orbbec Astra S 3D Camera %\footnote{\url{https://www.amazon.com/dp/B07484SMB8}} 
mounted on its frame, which captures 640$\times$480 images, and a Seeed Studio ReSpeaker microphone  %\footnote{\url{https://www.amazon.com/dp/B07ZGZSBS4}} 
placed on its workstation.
We recorded data from 14 and 11 sensor modalities for {\it Baxter} and {\it UR5}, respectively.
% For more dataset details, such as sampling rate, please see: {\small\url{https://go.tufts.edu/Baxter_UR5_95_Objects}}.
For more dataset details, such as sampling rate, please see: {\small\url{https://github.com/gtatiya/Implicit-Knowledge-Transfer}}.

\subsubsection{Exploratory Behaviors and Objects}

Both robots perform 8 behaviors: {\it look}, {\it grasp}, {\it pick}, {\it hold}, {\it shake}, {\it lower}, {\it drop}, and {\it push} (Fig. \ref{fig:System_Overview}C).
We chose these diverse behaviors because they can capture various object properties.
{\it Look} is a non-interactive behavior in which robots record visual modalities ({\it RGB}, {\it Depth}, and {\it Point-Cloud}) from their head camera.
All other behaviors are interactive, encoded as robot joint-angle trajectories.
% For all behaviors, {\it Point-Cloud} data was only recorded for the first 5 frames.
For all behaviors, {\it Point-Cloud} was recorded for the first 5 frames.
% 8-ounce cylindrical plastic containers
Both robots explore 95 objects (cylindrical containers) that vary in 5 colors ({\it blue}, {\it green}, {\it red}, {\it white}, and {\it yellow}), 6 contents ({\it wooden buttons}, {\it plastic dices}, {\it glass marbles}, {\it nuts} \& {\it bolts}, {\it pasta}, and {\it rice}), and 4 weights ({\it empty}, {\it 50g}, {\it 100g}, and {\it 150g}) shown in Fig. \ref{fig:System_Overview}D. %\footnote{\url{https://www.amazon.com/dp/B07WTZDJ1R}}.
There are 90 objects with contents (5 colors x 3 weights x 6 contents) and 5 objects without any content that only vary by 5 colors.

\subsubsection{Data Collection}

While recording sensor data, robots perform all 8 behaviors in a sequence on the 95 objects, in round-robin fashion, %. Once an object was explored, the same object was not explored again until all the objects were explored, 
to minimize any transient noise effects after a single trial on an object.
Both robots perform 5 such trials on each object, resulting in 7,600 interactions, overall.% (2 robots $\times$ 8 behaviors $\times$ 95 objects $\times$ 5 trials).

\subsubsection{Feature Extraction}

\begin{figure*}[!ht]
\centering
\includegraphics[width=0.9\linewidth]{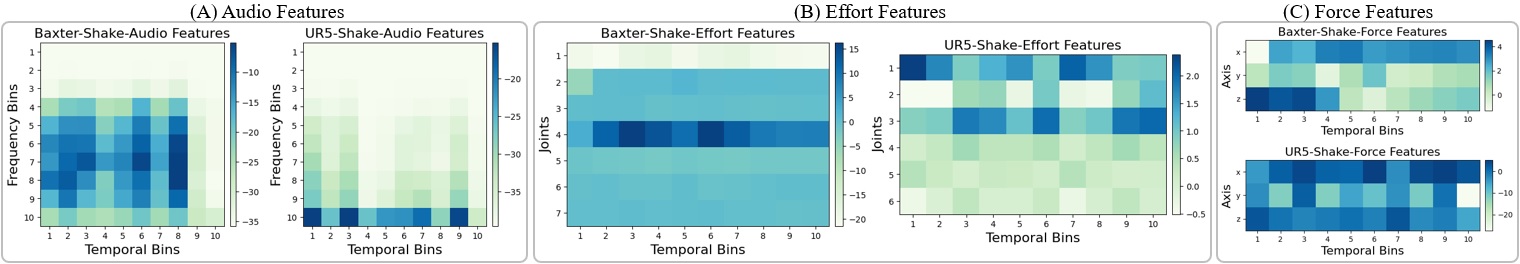}
\caption{\small Examples of (A) {\it audio}, (B) {\it effort} and (C) {\it force} features when {\it Baxter} and {\it UR5} perform {\it shake} on a {\it blue-marbles-150g} object.}
\label{fig:Features}
\vspace{-0.5cm}
\end{figure*}

We used all interactive\cut{\footnote{Experiments with {\it look} histogram features were performed, but no improvements were observed, indicating that vision alone is insufficient.}} behaviors in our experiments (i.e., all behaviors listed above except {\it look}). %For our experiments, we used all interactive behaviors (i.e., all behaviors listed above except {\it look}).
We used audio, effort at the robot's joints, and force at the robot's end-effector in our experiments, as they play crucial roles in the human somatosensory system for recognizing non-visual object properties.
For audio, we used librosa \cite{mcfee_librosa_2015} to generate mel-scaled spectrograms of the audio wave files recorded by robots with FFT window length of 1024, hop length of 512, and 60 mel-bands.
Then, a spectro-temporal histogram was computed by discretizing both time and frequency into 10 equally-spaced bins, where each bin consists of the mean of values in that bin.
Effort and force data were discretized into 10 equally-spaced temporal bins for joints and axes, respectively.
Thus, audio and force data are represented as 100 and 30 dimensional feature vectors, respectively.
For {\it Baxter} and {\it UR5}, {\it effort} data is represented as 70 and 60 dimensional feature vectors, respectively.
Fig. \ref{fig:Features} visualizes both robots' {\it audio}, {\it effort}, and {\it force} features when they perform {\it shake} behavior on a {\it blue-marbles-150g} object.

\begin{comment}
\begin{figure}
\centering
\includegraphics[width=7cm]{figs/Features.jpg}
\caption{\small Visualizations of {\it audio}, {\it effort} and {\it force} features (top to bottom) when \textbf{(left)} {\it Baxter} and \textbf{(right)} {\it UR5} perform the {\it shake} behavior on a {\it blue-marbles-150g} object.}
\label{fig:Features}
\end{figure}
\end{comment}

\subsubsection{Data Augmentation}

To improve model generalization, we increase the number of object trials through data augmentation: we compute each bin's mean and standard deviation in the discretized representation of all object trials and sample $k=5$ additional trials of each object.\cut{\footnote{Experiments with $k=10$ showed little additional improvement.}}
The rationale behind augmenting data by constraining on trials is to generate realistic data that is less likely to be impossible to produce in the real-world.
Furthermore, this data augmentation technique is independent of the downstream task and can be applied for both object-property and object-identity recognition.

\subsection{Evaluation}

We evaluated performance of the projection methods: 1) EDN projects source robot features to a target robot's feature space, and 2) KEMA projects individual robot features to a shared feature space.
To learn both projections, we evaluate two ways to build correspondence between source-target data pairs: 1) object identity-based pairs, wherein both source and target robots interact with the same object identity (e.g., {\it baxter-buttons-50g} and {\it ur5-buttons-50g}); and 2) object property-based pairs, wherein source and target robots interact with objects that share a property (e.g., {\it baxter-buttons-50g} and {\it ur5-dices-50g}, wherein weight is same and contents are different).
In both correspondence types, we use the same behavior and modality for both source and target robots. We consider 2 tasks: object property-recognition and object identity-recognition.
In property-recognition, the target robot must recognize content and weight of the object it interacts with; there are 7 content classes and 4 weight classes, including an empty class.
In object identity recognition, the target robot must recognize the specific object identity.

% \subsubsection{Object property recognition task} We train the target robot for the baseline condition using the data in its own feature space.
% We train the target robot using the projected features obtained by applying the projection approaches mentioned above for the transfer condition.
% In the training phase of the projection approach, we use $100\%$ of the source robot's data and increment the number of objects the target robot interacts with up to $80\%$.
% The remaining $20\%$ of the object is held-out for testing the target robot's performance.
% For both conditions, we performed a 10-fold cross-validation, in which we randomly sampled $80\%$ objects for training and used the rest $20\%$ objects for testing.
% To find the best performance of the task, we train the target robot using $100\%$ of the objects and evaluate on the test objects for each fold.

\subsubsection{Object Property Recognition Task}
As a baseline condition, we train the target robot using data in its own feature space.
For the transfer condition, we train the target robot using features obtained by applying the projections.
In training each projection, we use all 95 objects for the source robot and increment the number of objects the target robot interacts with, from 4 (for weight-recognition) and 7 (content-recognition), to 76 objects ($80\%$ of objects).
The remaining 19 objects ($20\%$ objects) are held-out for testing target robot performance.
% In both conditions, we performed 10-fold cross-validation: we randomly-sampled 76 objects for incremental training and used remaining 19 for testing.
We randomly-sampled 76 objects for incremental training and used remaining 19 for testing; we repeated this process 10 times, in both conditions.
For best target robot performance in the baseline condition, we train using all 95 objects and evaluate on test objects in each fold.
In all cases, we use all 5 trials of each object.

\subsubsection{Object Identity Recognition Task}
The baseline and transfer conditions of the object identity recognition task are the same as in the property recognition task.
We evaluated the target robot's performance to recognize 12 randomly-sampled objects from the 95 objects.
When training each projection method, we used all 5 trials of each object for the source robot and increment the number of trial per object from 1 to 4 ($80\%$ trials) for the target robot.
The remaining 1 trial ($20\%$ trials) of each object is held-out for testing the target robot's performance.
For both conditions, we performed 5-fold cross-validation such that each trial of all 12 objects is included in the test set, once.
For best target robot performance in the baseline condition, %of the target robot when it learns the task on its own feature space, 
we train using all 5 trials of all 12 objects and evaluate on the test trial of each object for each fold.
The process of selecting 12 objects, and performing 5-fold cross-validation for both conditions is repeated 10 times to compute performance statistics.

\subsubsection{Evaluation Metrics} We used two metrics to evaluate the target robot's recognition performance.
First, we consider accuracy $A = \frac{\text{correct \; predictions}}{\text{total predictions}} \%$; the second metric is the accuracy delta ($\Delta A$), which measures the drop in accuracy due to using projected features (obtained by interacting with \textit{fewer} objects) versus using the target robot's own features (obtained by interacting with \textit{all} objects).
We compute mean accuracy delta of the least $m$ number of object interactions in our experiments, defined as: %
$\text{m}\Delta A = \frac{1}{m} \sum_{j=1}^m (A_{all} - A_{projected}^{j}) \%$,
\begin{comment}
\begin{equation}
% \begin{displaymath}
A\Delta = \frac{1}{m} \sum_{j=1}^m (A_{all} - A_{projected}^{j}) \%
% \end{displaymath}
\end{equation} % Where $A_{all}$ and $A_{projected}$ are the accuracies obtained when using real and projected features, respectively, and $m = 1, ..., 10$.
\end{comment}
where $A_{all}$ is the accuracy obtained using $100\%$ of the target robot's data, $A_{projected}$ is the accuracy obtained using projected features, and $m=10$ for object property recognition, and $m=4$ for object identity recognition.
%To report both metrics' results, 
For both metrics, we use recognition accuracy computed as a weighted combination of all the behaviors and modalities used, based on their performance on the training data.

\section{Results}

\begin{figure}
\centering
\includegraphics[width=0.9\linewidth]{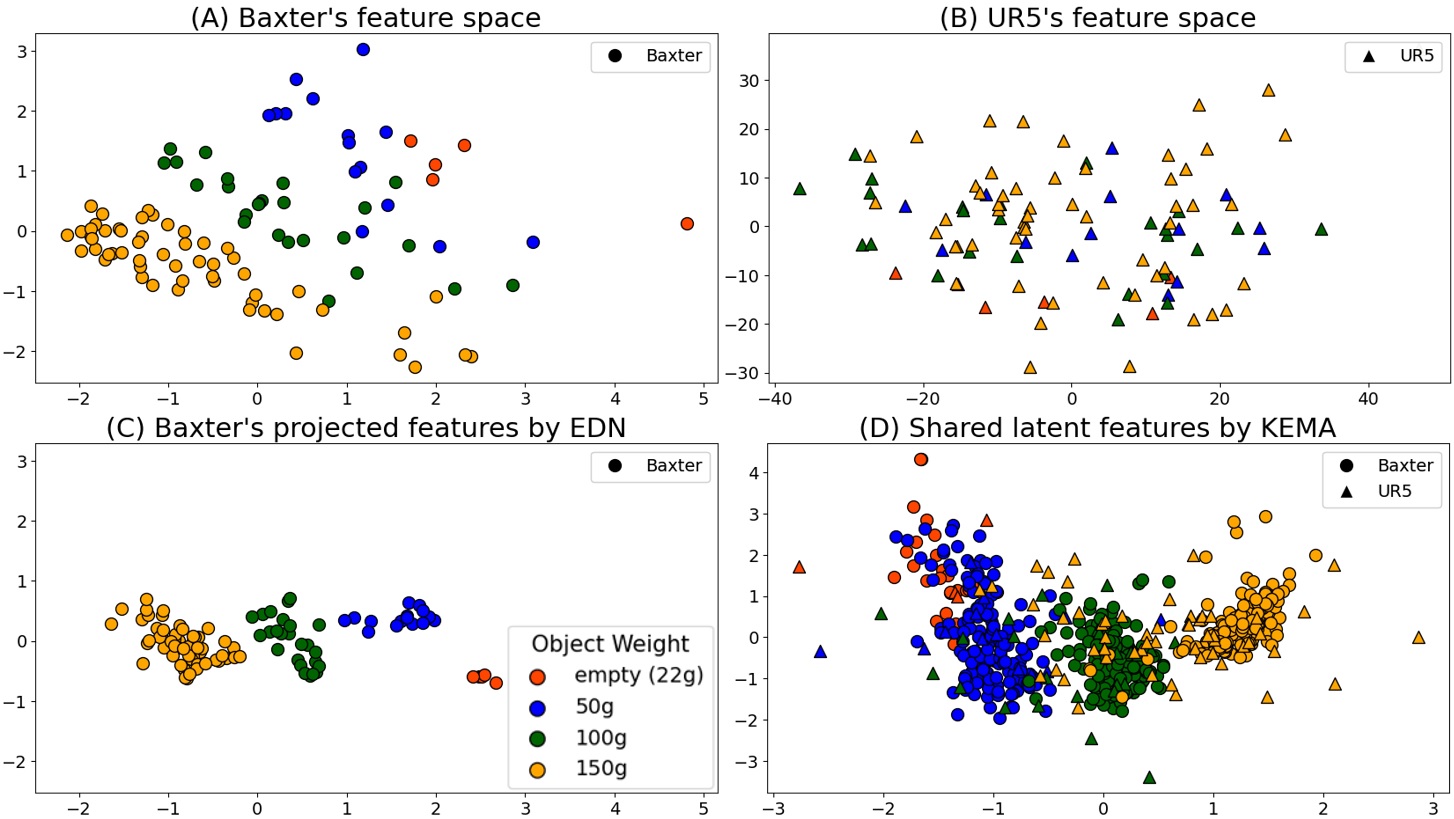}
\caption{\small Original sensory features of (A) {\it Baxter} and (B) {\it UR5} for {\it pick-force} performed on 20 objects in 2D space, and (C) the projected features from {\it UR5-pick-force} to {\it Baxter-pick-force} projection using EDN, and (D) first 2 dimensions of corresponding features in the shared latent feature space generated using KEMA.\vspace{-0.5cm}}
\label{fig:IE}
\end{figure}

\parai{Illustrative Example.}
Consider the case where a source robot ({\it Baxter}) and a target robot ({\it UR5}) perform {\it pick} behavior and record force signal. % at their end-point.
{\it Baxter} interacts with all 95 objects, and {\it UR5} interacts with only 20 objects;
both robots perform 5 trials on each object.
We use Principal Component Analysis (PCA) to visualize the robots' feature spaces (Fig. \ref{fig:IE}A and \ref{fig:IE}B) and plot object weights with different colors.
In Fig. \ref{fig:IE}A we only plot {\it Baxter's} features of the common 20 objects, for comparison to original and projected features shown in Fig. \ref{fig:IE}C.
We project {\it UR5-pick-force} to {\it Baxter-pick-force}, via EDN with object identity-based correspondences, and visualize with PCA in Fig. \ref{fig:IE}C.
Compared to {\it Baxter}'s space (Fig. \ref{fig:IE}A), projected features are more tightly clustered for different weights.
We also generate the shared latent features using KEMA with object identity-based correspondences.
% We use PCA to visualize the latent features in Fig. \ref{fig:IE}D: data collected by both robots of 4 different weights are clustered together, indicating both robots' data distribution is aligned efficiently.
We plot first 2 dimensions of latent features in Fig. \ref{fig:IE}D: data collected by both robots of 4 different weights are clustered together, indicating both robots' data distribution is aligned efficiently.

Consider another case where {\it UR5} interacts with one object of each weight 5 times and learns to recognize the object's weight using 20 examples (4 weights $\times$ 5 trials).
The mean accuracy computed over 10 folds using these 20 examples is $22.31 \pm 8.05$.
This learning process is the same as in our baseline condition, where the robot learns using its own features.
Now, we additionally use the 5 trials with data augmentation and train {\it UR5} to recognize the object's weight using 40 examples: 4 weights $\times$ (5 real trials + 5 augmented).
The mean accuracy computed over 10 folds using these real and augmented data is $28.21 \pm 6.09$; 
the increased accuracy shows that using data augmentation improves recognition performance.
Since, we consistently observed improvements from augmentation, we only report the performance of our baseline and transfer conditions using augmentation.

\parai{Object Property Recognition Results.}
%
% For transferring object property knowledge from the source to the target robot, we evaluated both EDN and KEMA approaches by building correspondences based on object-identity and object-property.
For the object property-recognition task, we evaluated both projection methods by building correspondences based on object-identity and object-property.
% We assume that the source robot has interacted with all 95 objects to evaluate both approaches.
% We incrementally varied the number of objects the target robot interacts with for training from 4 for weight recognition and 7 for content recognition to 76 objects ($80\%$ of objects).
% To test the target robot's performance, we used the held-out 19 objects ($20\%$ of objects).
% For both robots, we use all 5 trials of each object they interacted with.
For EDN, we built object-identity correspondences by mapping each source robot's object trial to all the target robot's trials of that object. We built object-property correspondences by mapping each source robot's object with a property of the recognition task to all the target robot's objects with that property.
For example, for the weight recognition task, a {\it 50g} object interacted by the source robot will be mapped to all the {\it 50g} objects interacted by the target robot.
For KEMA, we build the manifold alignment using all 95 objects of the source robot and incrementally vary the number of objects the target robot interacts with, for both object-identity and object-property correspondences.

% We compare the recognition accuracy of the baseline condition, where the target robot learns to recognize object properties using only its own features, and the transfer condition, where the target robot learns to recognize object properties using the projected features using EDN and KEMA.
% In both conditions, the recognition accuracy is computed by performing a weighted combination of all the behaviors and modalities based on their performance on the training examples.
% For comparison with the best possible performance of the task, we compute the recognition accuracy when the target robot uses all 95 objects for training.

\begin{figure}
\centering
\includegraphics[width=0.9\linewidth]
{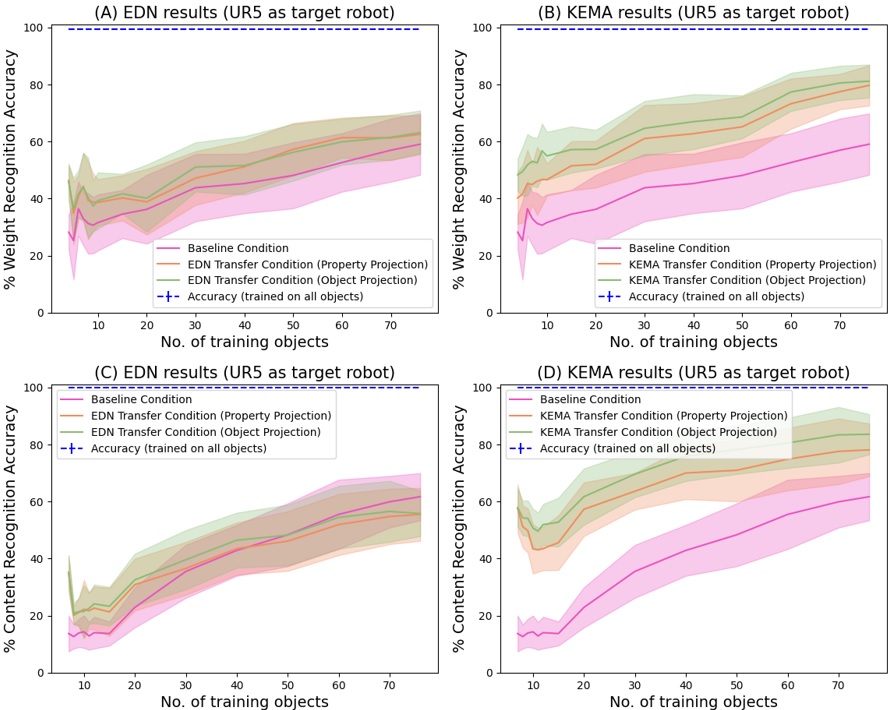}
\caption{\small Accuracy results of the baseline and transfer conditions, EDN (\textbf{left}) and KEMA (\textbf{right}), on the weight (\textbf{top}) and content (\textbf{bottom}) recognition tasks, for {\it Baxter} (source) and {\it UR5} (target).\vspace{-0.2cm}}
\label{fig:UR5_object_property_recognition_results}
\end{figure}

Fig. \ref{fig:UR5_object_property_recognition_results} shows results of EDN and KEMA on the weight-  and content-recognition tasks, where {\it Baxter} is the source robot and {\it UR5} is the target robot: all transfer conditions for both approaches perform better than the baseline condition when the target robot interacts with fewer objects.
% The transfer conditions perform consistently better than the baseline for the weight recognition task.
% However, for the content recognition task, as the target robot interacts with more objects, EDN performs comparable to the baseline condition, and KEMA suffers from the negative transfer.
As the target robot interacts with more objects, KEMA still performs better than baseline condition, and EDN performs comparable to baseline condition.
% This could be because the audio modality is crucial for content recognition and the audio signal transferred by {\it Baxter} was noisy due to its loud motors.
% EDN approach was able to take {\it Baxter's} noisy audio signal and generated denoised {\it UR5} audio signal.
% However, KEMA could not align both robots' audio signals, primarily when the target robot interacted with more objects.
Overall, results indicate that the proposed knowledge transfer methods can boost target robot performance, notably when it has limited time to learn tasks and cannot interact with many objects.
We also experimented with {\it UR5} as the source robot and {\it Baxter} as the target, and observed a similar performance boost with transfer. % (Fig. \ref{fig:Baxter_object_property_recognition_results}).

\begin{figure}
\centering
\includegraphics[width=0.9\linewidth]
{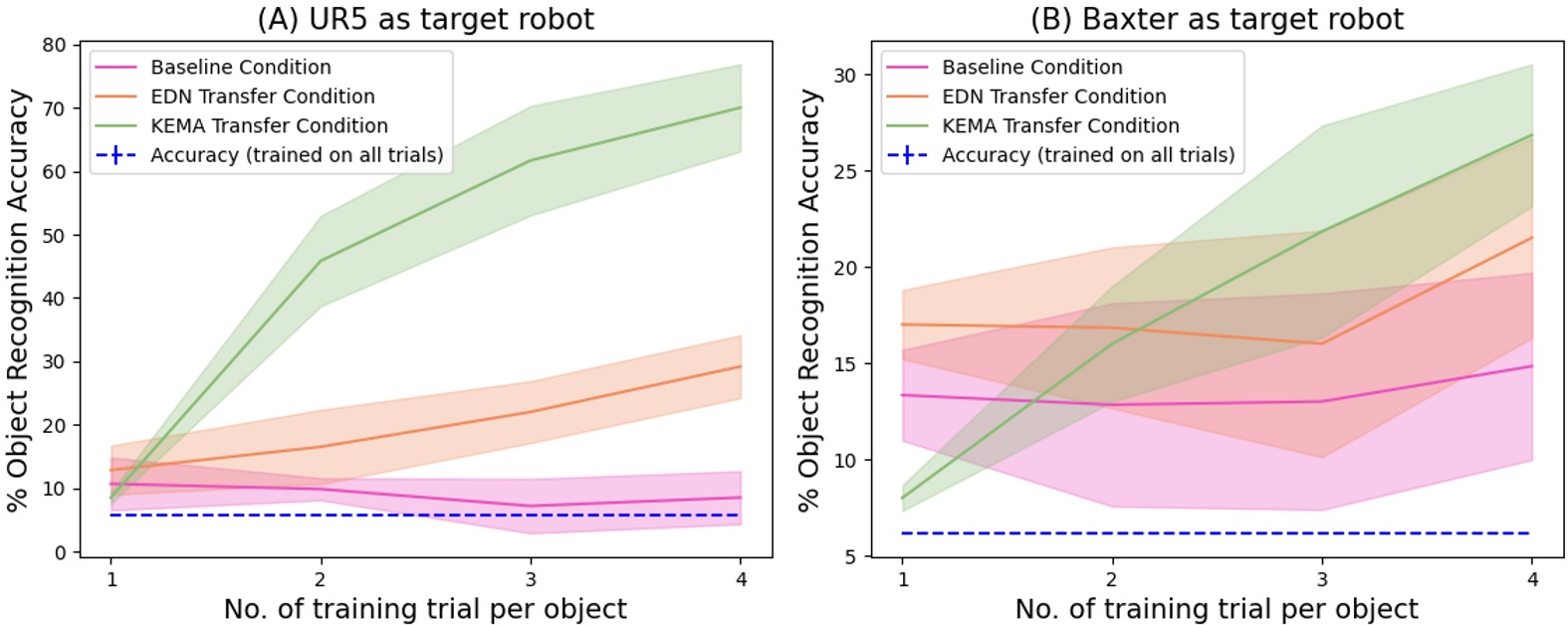}
\caption{\small Accuracy results of the baseline and transfer conditions on the object identity recognition tasks.\vspace{-0.5cm}}
\label{fig:object_identity_recognition_results}
\end{figure}

\begin{comment}
\begin{figure}
\centering
\includegraphics[width=8.6cm]{figs/Baxter_object_property_recognition_results_aug.jpg}
\caption{\small Accuracy results of the baseline and transfer conditions, (\textbf{left}) EDN and (\textbf{right}) KEMA, on the (\textbf{top}) weight and (\textbf{bottom}) content recognition tasks, where {\it UR5} serves as the source robots, and {\it Baxter} serves as the target robot.}
\label{fig:Baxter_object_property_recognition_results}
\end{figure}
\end{comment}

Table \ref{tab:mean_accuracy_delta} shows mean accuracy delta ($\text{m}\Delta A$) results of both methods and both correspondence types.
% A lower mean accuracy delta means better performance as it is closer to the best possible task performance.
Lower $\text{m}\Delta A$ means better performance, i.e., closer to the case where the target robot is trained using its own data with all objects.
For KEMA, object-identity correspondence yields better performance; for EDN, both correspondences perform comparably.
These findings indicate that object-identity correspondence builds better alignment for projecting features into the shared latent space than object property correspondence, though the correspondences yield comparable performance for projecting features into the target feature space.
KEMA outperforms EDN in all cases, showing KEMA as more efficient in transferring implicit object property knowledge across robots.
% Note that for this case transferring information in the audio signal is crucial, and Baxter's audio signal was noise.
% These results indicate that when we transfer knowledge free from noise, projecting features from robots into a shared latent space produces a better representation of the sensory signal than projection features into the target robot's feature space.
% In addition, when we transfer a noisy signal, generating features in the target robot's feature space produces a better representation of the sensory signal.

\parai{Object Identity Recognition Results.}
%
% To transfer object identity knowledge from the source to the target robot, we evaluated EDN and KEMA approaches by building correspondences based on object-identity.
For the object identity recognition task, we evaluated EDN and KEMA approaches by building correspondences based on object-identity.
We build the object-identity correspondences in the same manner as for the object property recognition task.
% We emphasize that identifying specific objects in our dataset is challenging because each object varies by color, content, and weight, and we only use non-visual modalities in our experiments.
\begin{table}[!h]
\centering
\caption{\small Mean accuracy delta ($\text{m}\Delta A$) results of EDN and KEMA for object identity-based and property-based correspondences.}
\label{tab:mean_accuracy_delta}
\scriptsize
\resizebox{0.9\columnwidth}{!}{
\begin{tabular}{p{0.40\linewidth}p{0.1\linewidth}p{0.1\linewidth}p{0.1\linewidth}p{0.1\linewidth}}
\toprule
 & \multicolumn{2}{c}{\textbf{\textsc{{\it \textbf{UR5}}}}} &  \multicolumn{2}{c}{\textbf{\textsc{{\it \textbf{Baxter}}}}} \\
\cmidrule(r){2-3} \cmidrule(r){4-5} \textbf{Method (correspondence)} & Weight & Content & Weight & Content \\
\midrule 
{EDN (object-identity)} & 57.72 & 71.26 & 31.91 & 32.88 \\
{EDN (object-property)} & 58.57 & 72.54 & 30.49 & 32.78 \\
{KEMA (object-identity)} & \textbf{44.88} & \textbf{42.17} & \textbf{28.84} & \textbf{19.25} \\
{KEMA (object-property)} & 51.88 & 47.53 & 34.88 & 22.60 \\
\bottomrule
\end{tabular}
}
\vspace{-0.3cm}
\end{table}
\begin{table}[!h]
\centering
\caption{\small Mean accuracy delta ($\text{m}\Delta A$) results of EDN and KEMA on the object identity recognition tasks.}
\label{tab:object_identity_recognition_results}
\scriptsize
\resizebox{0.9\columnwidth}{!}{
\begin{tabular}{p{0.35\linewidth}p{0.15\linewidth}p{0.1\linewidth}}
\toprule
\textbf{Method (correspondence)} & {\it \textbf{UR5}} & {\it \textbf{Baxter}} \\
\midrule
EDN (object-identity) & -14.29 & -11.67 \\
KEMA (object-identity) & \textbf{-40.67} & \textbf{-12.00} \\
\bottomrule
\end{tabular}
}
\vspace{-0.4cm}
\end{table}
We emphasize that identifying specific objects in our dataset is a challenging task. For example, if two objects have the same weight but different contents, it would be very crucial to listen to the audio signal produced while performing behaviors, as the force signal would not be helpful to distinguish those objects. Thus, we used 12 randomly-sampled objects with unique weight and content for the object identity recognition task.

Fig. \ref{fig:object_identity_recognition_results} shows the accuracy results, and Table \ref{tab:object_identity_recognition_results} shows the mean accuracy delta results of both approaches on object-identity correspondence.
Overall, both approaches perform better than the baseline condition, and KEMA performs significantly better than EDN.
% These results indicate that for identifying specific objects, features in the target feature space contain more helpful information than the shared latent space.
These results indicate that features in the shared latent space contain more helpful information for identifying specific objects than the target feature space.
% Furthermore, audio is crucial for object identity recognition, and EDN denoises sound signal from {\it Baxter} and projects to the {\it UR5's} feature space; thus we observe better performance when {\it UR5} is serving as the target robot as compared to when {\it Baxter} is serving as the target robot.
Negative values in Table \ref{tab:object_identity_recognition_results} show that using projected features for training the target robot leads to better performance than using $100\%$ of the target robot's own features.
These results indicate that using projected features from the source robot helps the target robot to learn a recognition model that generalizes better for object identity recognition.
In addition, our baseline condition also performs better than using $100\%$ of the target robot's own features, indicating that the data augmentation technique we applied improves the generalization of the recognition models.

%%%%%%%%%%%%%%%%%%%%%%%%%%%%%%%%%%%%%%%%%%%%%%%%%%%%%%%%%%%%%%%%%%%%%%%%%%%%%%%%
\section{Conclusion and Future Work}

For a robot to learn about implicit object properties, it must perform object exploration while processing various non-visual modalities. This process is costly across multiple robots as object feature representations are unique to a robot's morphology. We proposed a framework for transferring implicit object property knowledge across heterogeneous robots and evaluated two projection methods, on two interactive perception tasks; results showed that learning on a target robot is accelerated through transfer from source robot, even if it explores fewer objects.
%For robots to learn about object properties, they must explore objects and process received non-visual sensor signals; these signals contain crucial information that cannot be extracted by vision alone. Interactive object exploration is especially costly when multiple robots must learn interactive perception-based tasks, as object feature representations are unique to a robot's morphology. This paper proposes a framework to transfer implicit knowledge about object properties from a more-experienced source robot to a newly-deployed target robot: we evaluate two projection functions, encoder-decoder network (EDN) and kernel manifold alignment (KEMA), on two interactive perception tasks, object-property recognition and object-identity recognition. Results indicate that our framework can accelerate the target robot's learning, primarily when it explores fewer objects.
% In future work, we will investigate how best to (i) select source and target sensorimotor contexts (i.e., modality and behavior) and (ii) select the set of objects explored by robots for learning projections more efficiently. %We also plan to intelligently select the set of objects explored by robots to learn projections with less object exploration. 
% Moreover, we will leverage our proposed framework to transfer more object properties, e.g., shape, material, and stiffness.
%
Although our framework expedites learning on the less experienced target robot, there are some limitations.
We encoded different behaviors in robots for object exploration.
In future work, we plan to enable robots to learn behaviors to extract different object properties, autonomously.
Moreover, we assumed that both source and target robots explored objects using the same sensorimotor context; thus, we used this same context while learning the projections.
We plan to select sensorimotor contexts for learning projections more efficiently.
Furthermore, we plan to automate the selection of objects to be explored, to learn the projection faster. % with lesser object exploration.
%Finally, another promising direction of future work is to develop a framework for a scenario where more than two robots explore objects with additional properties such as shape, size, material, and stiffness.
Finally, we envision a scenario where more than two robots explore objects with additional properties, e.g., shape, size, material, and stiffness.

\clearpage

\bibliographystyle{IEEEtran}
\bibliography{references}

\begin{thebibliography}{10}
\providecommand{\url}[1]{#1}
\csname url@samestyle\endcsname
\providecommand{\newblock}{\relax}
\providecommand{\bibinfo}[2]{#2}
\providecommand{\BIBentrySTDinterwordspacing}{\spaceskip=0pt\relax}
\providecommand{\BIBentryALTinterwordstretchfactor}{4}
\providecommand{\BIBentryALTinterwordspacing}{\spaceskip=\fontdimen2\font plus
\BIBentryALTinterwordstretchfactor\fontdimen3\font minus
  \fontdimen4\font\relax}
\providecommand{\BIBforeignlanguage}[2]{{%
\expandafter\ifx\csname l@#1\endcsname\relax
\typeout{** WARNING: IEEEtran.bst: No hyphenation pattern has been}%
\typeout{** loaded for the language `#1'. Using the pattern for}%
\typeout{** the default language instead.}%
\else
\language=\csname l@#1\endcsname
\fi
#2}}
\providecommand{\BIBdecl}{\relax}
\BIBdecl

\bibitem{thesen_Neuroimaging_2004}
\BIBentryALTinterwordspacing
T.~Thesen, J.~F. Vibell, G.~A. Calvert, and R.~A. Osterbauer, ``Neuroimaging of
  multisensory processing in vision, audition, touch, and olfaction,''
  \emph{Cognitive processing}, vol.~5, no.~2, pp. 84--93, 2004. [Online].
  Available: \url{https://doi.org/10.1007/s10339-004-0012-4}
\BIBentrySTDinterwordspacing

\bibitem{lederman1987hand}
S.~J. Lederman and R.~L. Klatzky, ``Hand movements: A window into haptic object
  recognition,'' \emph{Cognitive psychology}, vol.~19, no.~3, pp. 342--368,
  1987.

\bibitem{wilcox2007multisensory}
T.~Wilcox, R.~Woods, C.~Chapa, and S.~McCurry, ``Multisensory exploration and
  object individuation in infancy.'' \emph{Developmental psychology}, vol.~43,
  no.~2, p. 479, 2007.

\bibitem{ernst2004merging}
M.~O. Ernst and H.~H. B{\"u}lthoff, ``Merging the senses into a robust
  percept,'' \emph{Trends in cognitive sciences}, vol.~8, no.~4, pp. 162--169,
  2004.

\bibitem{gibson1988exploratory}
E.~J. Gibson, ``Exploratory behavior in the development of perceiving, acting,
  and the acquiring of knowledge,'' \emph{Annual review of psychology},
  vol.~39, no.~1, pp. 1--42, 1988.

\bibitem{chen2021framework}
X.~Chen, R.~Hosseini, K.~Panetta, and J.~Sinapov, ``A framework for
  multisensory foresight for embodied agents,'' in \emph{2021 IEEE
  International Conference on Robotics and Automation (ICRA)}.\hskip 1em plus
  0.5em minus 0.4em\relax IEEE, 2021, pp. 10\,927--10\,933.

\bibitem{mccarthy1989artificial}
J.~McCarthy, ``Artificial intelligence, logic and formalizing common sense,''
  in \emph{Philosophical logic and artificial intelligence}.\hskip 1em plus
  0.5em minus 0.4em\relax Springer, 1989, pp. 161--190.

\bibitem{erickson2020multimodal}
Z.~Erickson, E.~Xing, B.~Srirangam, S.~Chernova, and C.~C. Kemp, ``Multimodal
  material classification for robots using spectroscopy and high resolution
  texture imaging,'' in \emph{2020 IEEE/RSJ International Conference on
  Intelligent Robots and Systems (IROS)}.\hskip 1em plus 0.5em minus
  0.4em\relax IEEE, 2020, pp. 10\,452--10\,459.

\bibitem{huang2022understanding}
H.-J. Huang, X.~Guo, and W.~Yuan, ``Understanding dynamic tactile sensing for
  liquid property estimation,'' in \emph{Robotics Science and System}, 2022.

\bibitem{tatiya2019deep}
\BIBentryALTinterwordspacing
G.~Tatiya and J.~Sinapov, ``Deep multi-sensory object category recognition
  using interactive behavioral exploration,'' in \emph{International Conference
  on Robotics and Automation (ICRA), Montreal, QC, Canada, May 20-24,
  2019}.\hskip 1em plus 0.5em minus 0.4em\relax IEEE, 2019, pp. 7872--7878.
  [Online]. Available: \url{https://doi.org/10.1109/ICRA.2019.8794095}
\BIBentrySTDinterwordspacing

\bibitem{thomason2017opportunistic}
J.~Thomason, A.~Padmakumar, J.~Sinapov, J.~Hart, P.~Stone, and R.~J. Mooney,
  ``Opportunistic active learning for grounding natural language
  descriptions,'' in \emph{Conference on Robot Learning}.\hskip 1em plus 0.5em
  minus 0.4em\relax PMLR, 2017, pp. 67--76.

\bibitem{malinovska2022connectionist}
K.~Malinovsk{\'a}, I.~Farka{\v{s}}, J.~Harvanov{\'a}, and M.~Hoffmann, ``A
  connectionist model of associating proprioceptive and tactile modalities in a
  humanoid robot,'' in \emph{2022 IEEE International Conference on Development
  and Learning (ICDL)}.\hskip 1em plus 0.5em minus 0.4em\relax IEEE, 2022, pp.
  336--342.

\bibitem{wei2021multimodal}
J.~Wei, S.~Cui, J.~Hu, P.~Hao, S.~Wang, and Z.~Lou, ``Multimodal unknown
  surface material classification and its application to physical reasoning,''
  \emph{IEEE Transactions on Industrial Informatics}, vol.~18, no.~7, pp.
  4406--4416, 2021.

\bibitem{liu2022texture}
Y.~Liu, S.~Lu, and H.~Culbertson, ``Texture classification by audio-tactile
  crossmodal congruence,'' in \emph{2022 IEEE Haptics Symposium
  (HAPTICS)}.\hskip 1em plus 0.5em minus 0.4em\relax IEEE, 2022, pp. 1--7.

\bibitem{francis2022core}
J.~Francis, N.~Kitamura, F.~Labelle, X.~Lu, I.~Navarro, and J.~Oh, ``Core
  challenges in embodied vision-language planning,'' \emph{Journal of
  Artificial Intelligence Research}, vol.~74, pp. 459--515, 2022.

\bibitem{badrinarayanan2017segnet}
V.~Badrinarayanan, A.~Kendall, and R.~Cipolla, ``Segnet: A deep convolutional
  encoder-decoder architecture for image segmentation,'' \emph{IEEE
  transactions on pattern analysis and machine intelligence}, vol.~39, no.~12,
  pp. 2481--2495, 2017.

\bibitem{tatiya2020haptic}
\BIBentryALTinterwordspacing
G.~Tatiya, Y.~Shukla, M.~Edegware, and J.~Sinapov, ``Haptic knowledge transfer
  between heterogeneous robots using kernel manifold alignment,'' in
  \emph{{IEEE/RSJ} International Conference on Intelligent Robots and Systems
  {IROS}, Virtual Event, Las Vegas, NV, USA, October 25-29, 2020}.\hskip 1em
  plus 0.5em minus 0.4em\relax {IEEE}, 2020, pp. 5358--5363. [Online].
  Available: \url{https://doi.org/10.1109/IROS45743.2020.9340770}
\BIBentrySTDinterwordspacing

\bibitem{liu2018transferable}
Y.~Liu, Z.~Lu, J.~Li, C.~Yao, and Y.~Deng, ``Transferable feature
  representation for visible-to-infrared cross-dataset human action
  recognition,'' \emph{Complexity}, vol. 2018, 2018.

\bibitem{wang2011heterogeneous}
C.~Wang and S.~Mahadevan, ``Heterogeneous domain adaptation using manifold
  alignment,'' in \emph{Twenty-second international joint conference on
  artificial intelligence}, 2011.

\bibitem{klatzky1992stages}
R.~L. Klatzky and S.~J. Lederman, ``Stages of manual exploration in haptic
  object identification,'' \emph{Perception \& psychophysics}, vol.~52, no.~6,
  pp. 661--670, 1992.

\bibitem{lederman1993extracting}
S.~J. Lederman and R.~L. Klatzky, ``Extracting object properties through haptic
  exploration,'' \emph{Acta psychologica}, vol.~84, no.~1, pp. 29--40, 1993.

\bibitem{di2002role}
A.~Di~Ferdinando, A.~M. Borghi, and D.~Parisi, ``The role of action in object
  categorization.'' in \emph{FLAIRS Conference}, 2002, pp. 138--142.

\bibitem{lacey2007vision}
S.~Lacey, C.~Campbell, and K.~Sathian, ``Vision and touch: multiple or
  multisensory representations of objects?'' \emph{Perception}, vol.~36,
  no.~10, pp. 1513--1521, 2007.

\bibitem{lacey2014visuo}
S.~Lacey and K.~Sathian, ``Visuo-haptic multisensory object recognition,
  categorization, and representation,'' \emph{Frontiers in psychology}, vol.~5,
  p. 730, 2014.

\bibitem{bohg2017interactive}
J.~Bohg, K.~Hausman, B.~Sankaran, O.~Brock, D.~Kragic, S.~Schaal, and G.~S.
  Sukhatme, ``Interactive perception: Leveraging action in perception and
  perception in action,'' \emph{IEEE Transactions on Robotics}, vol.~33, no.~6,
  pp. 1273--1291, 2017.

\bibitem{pastor2020bayesian}
F.~Pastor, J.~Garc{\'\i}a-Gonz{\'a}lez, J.~M. Gandarias, D.~Medina, P.~Closas,
  A.~J. Garc{\'\i}a-Cerezo, and J.~M. G{\'o}mez-de Gabriel, ``Bayesian and
  neural inference on lstm-based object recognition from tactile and
  kinesthetic information,'' \emph{IEEE Robotics and Automation Letters},
  vol.~6, no.~1, pp. 231--238, 2020.

\bibitem{sawhney2020playing}
A.~Sawhney, S.~Lee, K.~Zhang, M.~Veloso, and O.~Kroemer, ``Playing with food:
  Learning food item representations through interactive exploration,'' in
  \emph{International Symposium on Experimental Robotics}.\hskip 1em plus 0.5em
  minus 0.4em\relax Springer, 2020, pp. 309--322.

\bibitem{navarro2022visuo}
N.~Navarro-Guerrero, S.~Toprak, J.~Josifovski, and L.~Jamone, ``Visuo-haptic
  object perception for robots: An overview,'' \emph{arXiv preprint
  arXiv:2203.11544}, 2022.

\bibitem{li2020review}
Q.~Li, O.~Kroemer, Z.~Su, F.~F. Veiga, M.~Kaboli, and H.~J. Ritter, ``A review
  of tactile information: Perception and action through touch,'' \emph{IEEE
  Transactions on Robotics}, vol.~36, no.~6, pp. 1619--1634, 2020.

\bibitem{wang2022audio}
Y.~Wang, K.~Wang, Y.~Wang, D.~Guo, H.~Liu, and F.~Sun, ``Audio-visual grounding
  referring expression for robotic manipulation,'' in \emph{2022 International
  Conference on Robotics and Automation (ICRA)}.\hskip 1em plus 0.5em minus
  0.4em\relax IEEE, 2022, pp. 9258--9264.

\bibitem{sinapov_grounding_2014}
\BIBentryALTinterwordspacing
J.~Sinapov, C.~Schenck, K.~Staley, V.~Sukhoy, and A.~Stoytchev, ``Grounding
  semantic categories in behavioral interactions: Experiments with 100
  objects,'' \emph{Robotics and Autonomous Systems}, vol.~62, no.~5, pp.
  632--645, may 2014. [Online]. Available:
  \url{https://www.sciencedirect.com/science/article/pii/S092188901200190X}
\BIBentrySTDinterwordspacing

\bibitem{falco2019transfer}
P.~Falco, S.~Lu, C.~Natale, S.~Pirozzi, and D.~Lee, ``A transfer learning
  approach to cross-modal object recognition: from visual observation to
  robotic haptic exploration,'' \emph{IEEE Transactions on Robotics}, vol.~35,
  no.~4, pp. 987--998, 2019.

\bibitem{tatiya2019sensorimotor}
\BIBentryALTinterwordspacing
G.~Tatiya, R.~Hosseini, M.~C. Hughes, and J.~Sinapov, ``Sensorimotor
  cross-behavior knowledge transfer for grounded category recognition,'' in
  \emph{Joint IEEE 9th International Conference on Development and Learning and
  Epigenetic Robotics (ICDL-EpiRob), Oslo, Norway, August 19-22, 2019}.\hskip
  1em plus 0.5em minus 0.4em\relax {IEEE}, 2019, pp. 1--6. [Online]. Available:
  \url{https://doi.org/10.1109/DEVLRN.2019.8850715}
\BIBentrySTDinterwordspacing

\bibitem{tatiya2020framework}
\BIBentryALTinterwordspacing
G.~Tatiya, R.~Hosseini, M.~Hughes, and J.~Sinapov, ``A framework for
  sensorimotor cross-perception and cross-behavior knowledge transfer for
  object categorization,'' \emph{Frontiers in Robotics and AI}, vol.~7, p. 137,
  2020. [Online]. Available:
  \url{https://www.frontiersin.org/article/10.3389/frobt.2020.522141}
\BIBentrySTDinterwordspacing

\bibitem{Gao_2022_CVPR}
R.~Gao, Z.~Si, Y.-Y. Chang, S.~Clarke, J.~Bohg, L.~Fei-Fei, W.~Yuan, and J.~Wu,
  ``Objectfolder 2.0: A multisensory object dataset for sim2real transfer,'' in
  \emph{Proceedings of the IEEE/CVF Conference on Computer Vision and Pattern
  Recognition (CVPR)}, June 2022, pp. 10\,598--10\,608.

\bibitem{tuia_kernel_2016}
\BIBentryALTinterwordspacing
D.~Tuia and G.~Camps-Valls, ``Kernel manifold alignment for domain
  adaptation,'' \emph{PLOS ONE}, vol.~11, no.~2, pp. 1--25, feb 2016. [Online].
  Available: \url{https://dx.plos.org/10.1371/journal.pone.0148655}
\BIBentrySTDinterwordspacing

\bibitem{kingma_adam_2015}
\BIBentryALTinterwordspacing
D.~P. Kingma and J.~Ba, ``Adam: A method for stochastic optimization,'' in
  \emph{International Conference on Learning Representations (ICLR)}, San
  Diego, CA, USA, may 2015. [Online]. Available:
  \url{https://arxiv.org/abs/1412.6980}
\BIBentrySTDinterwordspacing

\bibitem{baxter_fig}
``Baxter figure,''
  \url{https://www.computerhistory.org/collections/catalog/102751979},
  accessed: 2022-09-10.

\bibitem{ur5_fig}
``Ur5 figure,'' \url{https://www.nonead.com/en/intelligence_content/9667.html},
  accessed: 2022-09-10.

\bibitem{mcfee_librosa_2015}
\BIBentryALTinterwordspacing
B.~McFee, C.~Raffel, D.~Liang, D.~Ellis, M.~McVicar, E.~Battenberg, and
  O.~Nieto, ``librosa: Audio and music signal analysis in python,'' in
  \emph{Python in Science Conference (SciPy)}, vol.~14, Austin, Texas, 2015,
  pp. 18--24. [Online]. Available:
  \url{https://conference.scipy.org/proceedings/scipy2015/brian_mcfee.html}
\BIBentrySTDinterwordspacing

\end{thebibliography}

\end{document}